\documentclass[letterpaper]{article} 
\usepackage{aaai2026}  
\usepackage{times}  
\usepackage{helvet}  
\usepackage{courier}  
\usepackage[hyphens]{url}  
\usepackage{graphicx} 
\usepackage{natbib}  
\usepackage{caption} 
\usepackage{algorithm}
\usepackage{algorithmic}

\usepackage{newfloat}
\usepackage{listings}
\DeclareCaptionStyle{ruled}{labelfont=normalfont,labelsep=colon,strut=off} 
\floatstyle{ruled}
\newfloat{listing}{tb}{lst}{}
\floatname{listing}{Listing}
\usepackage{booktabs}       
\usepackage{nicefrac}       
\usepackage{microtype}      
\usepackage{afterpage,amsfonts,tabularx,caption,pifont,xcolor}
\usepackage{multirow}
\usepackage{siunitx}
\usepackage{amsmath}
\usepackage{amssymb}
\usepackage[export]{adjustbox}
\usepackage{subcaption}
\usepackage{colortbl,tabularray,bm}
\newcommand{\model}{FLAG-4D}
\title{\model: Flow-Guided Local-Global Dual-Deformation Model for 4D Reconstruction}
\author{
    Guan Yuan Tan\textsuperscript{\rm 1}\equalcontrib, Ngoc Tuan Vu\textsuperscript{\rm 1}\equalcontrib, Arghya Pal\textsuperscript{\rm 1}, Sailaja Rajanala\textsuperscript{\rm 1}, Rapha\"{e}l C.-W. Phan\textsuperscript{\rm 1},  Mettu Srinivas\textsuperscript{\rm 2}, Chee-Ming Ting\textsuperscript{\rm 1}
}
\affiliations{
    \textsuperscript{\rm 1} Monash University\\
    \textsuperscript{\rm 2} National Institute of Technology Warangal\\
}

\begin{document}

\maketitle

\begin{abstract}
We introduce \model, a novel framework for generating novel views of dynamic scenes by reconstructing how 3D Gaussian primitives evolve through space and time. 
Existing methods typically rely on a single Multilayer Perceptron (MLP) to model temporal deformations, and they often struggle to capture complex point motions and fine-grained dynamic details consistently over time, especially from sparse input views. 
Our approach, \model, overcomes this by employing a dual-deformation network that dynamically warps a canonical set of 3D Gaussians over time into new positions and anisotropic shapes. 
This dual-deformation network consists of an Instantaneous Deformation Network (IDN) for modeling fine-grained, local deformations and a Global Motion Network (GMN) for capturing long-range dynamics, refined through mutual learning. 
To ensure these deformations are both accurate and temporally smooth, \model\ incorporates dense motion features from a pretrained optical flow backbone. We fuse these motion cues from adjacent timeframes and use a deformation-guided attention mechanism to align this flow information with the current state of each evolving 3D Gaussian.
Extensive experiments demonstrate that \model\ achieves higher-fidelity and more temporally coherent reconstructions with finer detail preservation than state-of-the-art methods. 
\end{abstract}

\begin{links}
    \link{Code}{https://github.com/tgy1221/FLAG-4D}
\end{links}

\section{Introduction}

4D reconstruction has become a crucial advancement for capturing and reconstructing dynamic real-world objects and scenes, incorporating the temporal dimension \( t \) within the traditional 3D spatial coordinates \( (x, y, z) \). This approach enables the modeling of continuous changes, movements, and deformations in objects and environments over time, an essential feature for applications in Augmented Reality (AR) and Virtual Reality (VR), where accurate motion capture enhances immersive experiences. Recent works in Neural Radiance Fields (NeRF) \cite{mildenhall2021nerf} and 3D Gaussian Splatting (3DGS) \cite{kerbl20233d} have been extended to enable 4D reconstruction by integrating time-dependent deformation networks to capture and predict scene changes across temporal frames \cite{das2024neural, katsumata2023efficient, lu20243d, wu20244d}.

There are prior works that tried to extend 3D models to 4D through per-frame optimization; however, these approaches are costly, not very accurate in predicting Gaussian parameters, and have limited ability to handle dynamic scene elements or adapt to continuous deformations in complex object geometries. Existing methods \cite{yang2024deformable, huang2024sc, liang2023gaufre, sun20243dgstream} mitigate the first two limitations by using mean Gaussian location. However, these approaches largely fail to address the third limitation due to a fundamental dilemma in their design. A common underlying strategy involves employing a Multilayer Perceptron (MLP) to model all temporal deformations. This creates an inherent tension: the network capacity required to capture intricate local details is in disagreement with the need for smooth, globally coherent motion. Consequently, the models tuned with local details often produce temporally inconsistent results, while models tuned with global details often tend to over-smooth the scene, missing the high-frequency dynamics. This results in reconstructions that lack temporal consistency or fail to preserve dynamic details. 

While the principle of decomposing motion into local and global components has been explored, prior works often treat them as independent or loosely coupled systems. We introduce FLAG-4D, a dual-network of synergistic specialists designed to resolve the tension between local detail and global consistency. Our key contribution is not merely on the decomposition itself, but the tightly integrated mechanism by which these components interact. The framework consists of: (1) an Instantaneous Deformation Network (IDN), which excels at forecasting fine-grained, highly localized deformations, which are crucial for preserving texture and subtle movements, and (2) a Global Motion Network (GMN), which is designed to capture broader, scene-level dynamics and ensure long-range temporal consistency. 

The tight coupling between these specialists is achieved by the GMN, which leverages our proposed Contextual Deformation Alignment (CDA) mechanism. This deformation-guided attention process introduces a dynamic, query-based approach where the IDN's ``look-ahead'' local predictions act as a query that probes the GMN. The GMN then provides tailored, globally-aware guidance to the final deformation by using its CDA module to selectively retrieve relevant context from rich optical flow embeddings. This architecture is the first to use local motion forecast to guide the selection of global information. This differs from the simple feature concatenation or indirect loss-based supervision. This unique CDA process enables the specialists to synergize: the IDN preserves intricate local details from ambiguous monocular cues, while the GMN promotes global coherence necessary to resolve large-scale motion ambiguities. 
\begin{figure*}[!t]
\centering
\includegraphics[width=0.8\textwidth]{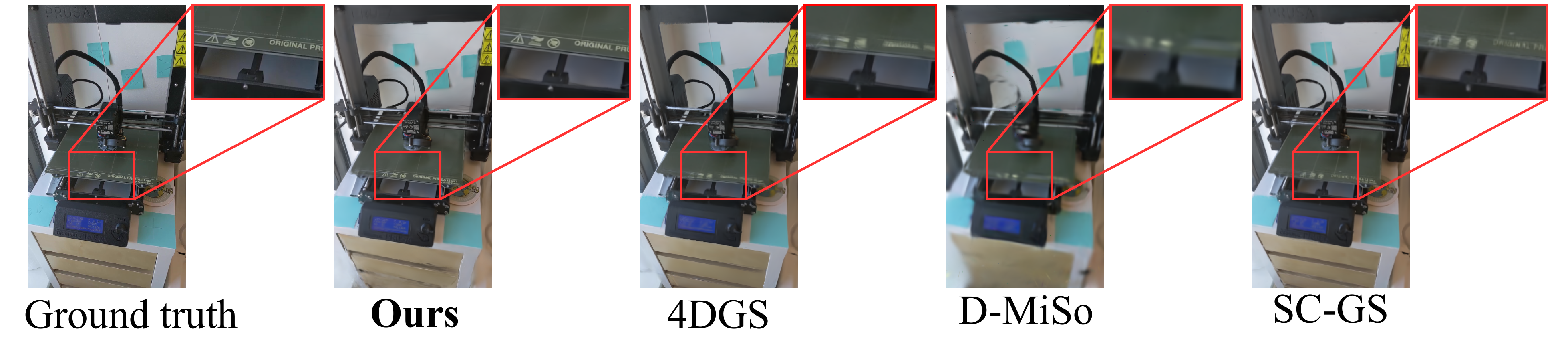}
    \caption{Visual Comparison of our method against very recent methods, such as 4DGS \cite{wu20244d}, SC-GS \cite{huang2024sc}, and D-MiSo \cite{waczynska2024d} on the HyperNeRF dataset. Our method demonstrates finer detail preservation across timesteps, particularly in the texture and edges of the zoomed-in regions. This results in a higher fidelity and coherent reconstruction across dynamic viewpoints compared to the baseline.}\label{fig:detailed-comparison}
\end{figure*}
Our contributions can be summarized as follows:
\begin{itemize}
    \item We propose a novel, tightly-coupled dual-network architecture of synergistic specialists: an Instantaneous Deformation Network (IDN) to model local details and a Global Motion Network (GMN) for global coherence.
    \item We introduce a novel query-based mechanism, Contextual Deformation Alignment (CDA), where a local deformation forecast is used as a dynamic query to selectively retrieve context from optical flow embeddings. 
    \item We introduce a synergistic mutual learning strategy to harmonize the specialist networks and enable them to learn complementary representations, enhancing temporal coherence and reconstruction quality.

\end{itemize}
\section{Related Work}
\label{sec:relatedwork}
\paragraph{Dynamic 3D Reconstruction}
Several foundation works in 4D reconstruction have advanced the use of Neural Radiance Fields (NeRF) \cite{mildenhall2021nerf} for dynamic scenes. Nerfies \cite{park2021nerfies}, Hypernerf \cite{park2021hypernerf}, D-NeRf \cite{pumarola2021d}, and DyNeRF \cite{li2022neural} have pioneered the application of deformation neural fields to extend the static NeRFs to capture dynamic scenes. However, NeRF-based approaches often suffer from high computational costs, limiting their real-time applicability \cite{mixtureofvolumetric,neuralresidual,wang2023mixed,park2021hypernerf}. Alternatives, such as explicit voxel grids \cite{fang2022fast} or plane-based factorizations \cite{shao2023tensor4d,fridovich2023k,cao2023hexplane}, have improved real-time flexibility by using MLP decoders for deformation. 

\paragraph{Dynamic 3D Gaussian Splatting}
Recently, Dynamic 3D Gaussian Splatting (3DGS) methods have gained significant traction due to their rendering speed and quality. \cite{wu20244d,duisterhof2023md} adapted plane-based encodings, or directly integrated temporal components into Gaussian representation \cite{li2024spacetime}. Many subsequent works rely on MLPs to predict Gaussian deformations over time \cite{yang2024deformable,yang2023real,sun20243dgstream}, with variations including sparse control points \cite{huang2024sc}, specialized per-Gaussian embeddings \cite{bae2024per}, and models that distinguish between static and deformable scenes \cite{liang2023gaufre}. Other approaches target temporal consistency specifically by employing tunable MLPs for varied motion patterns \cite{liu2024swings}, using time-calibrated inputs \cite{kim2024optimizing}, recovering deformations with motion bases \cite{kratimenos2025dynmf}, or explicit temporal modeling with polynomials and Fourier series to each Gaussian's trajectory \cite{lin2024gaussian}. These works solely depend on the mean Gaussian position and the current time step, while lacking relevant guidance, stressing on the deformation network to predict the deformation vectors in the next time step.

\paragraph{Optical Flow in 3D Dynamic Reconstruction}
Optical flow, refined by deep learning models \cite{flownet,pwcnet,raft,huang2022flowformer}, is increasingly used in dynamic 3D reconstruction to enhance temporal consistency and motion accuracy.
Optical flow has been applied to supervise 3D Gaussian movements \cite{flownet,pwcnet,raft,huang2022flowformer}, or as a component in the loss function to encourage smoother transitions \cite{liu2023robust,katsumata2025compact,zhu2024motiongs,gao2024gaussianflow}. In contrast to all these approaches, our work is the first to leverage the rich, pre-trained optical flow embedding as input to guide the deformation process directly. 

\section{Methodology}
\label{sec:methodology}
\begin{figure*}[ht] 
  \centering
  \includegraphics[width=0.88\textwidth]{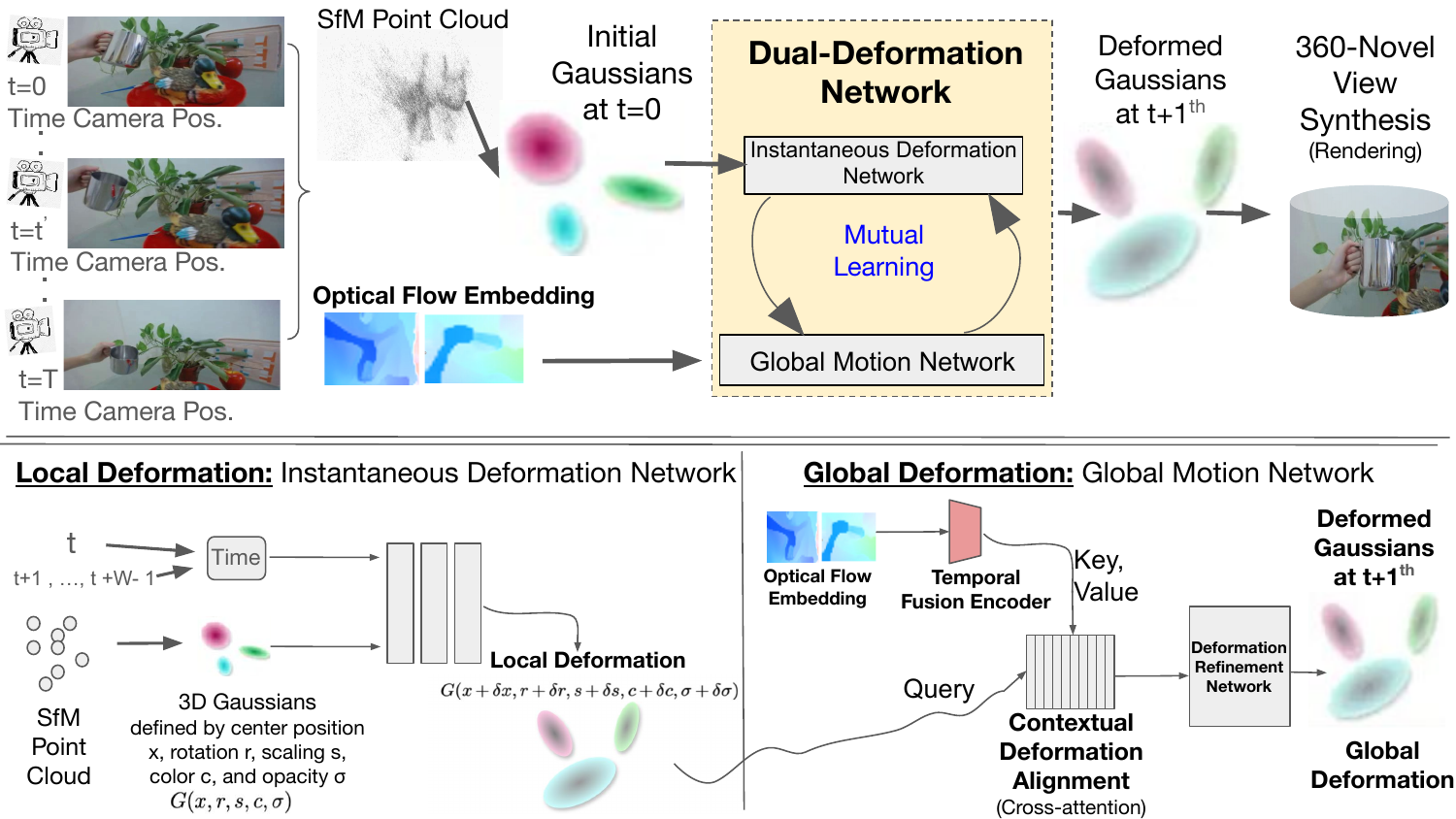}
    \caption{
    \textbf{\model\ Methodology:} Our dual-deformation framework for 4D reconstruction. \textbf{Top:} The overall pipeline: A monocular video sequence is used to generate an initial SfM point cloud, from which a canonical set of 3D Gaussians at $t=0$ is derived. The Dual-Deformation Network consists of an Instantaneous Deformation Network (IDN) and a Global Motion Network (GMN), which are trained synergistically through Mutual Learning. \textbf{Bottom Left:} The IDN processes the canonical Gaussians and a window of future-oriented time embeddings to produce a hypothesized local deformation. \textbf{Bottom Right:} The GMN integrates this local deformation hypothesis (as Query) with fused optical flow embeddings (as Key/Value) via a cross-attention mechanism, producing the final globally consistent deformation. 
    }\label{fig:bfd}
\end{figure*}
\paragraph{3D Gaussian Splatting}
3D Gaussian Splatting represents the scene with a set of explicit 3D Gaussians. Each primitive $k$ is characterized by: center location $\mu_k \in \mathbb{R}^3$, covariance matrix $\Sigma_k \in \mathbb{R}^{3\times3}$, opacity $o_k \in [0,1]$, and view-dependent color $c_k$, modeled with spherical harmonics coefficient $sh_k$. The covariance matrix $\Sigma_k$ can be decomposed as $\Sigma_k = R_kS_k{S_k}^T{R_k}^T$, where rotation matrix $R_k$ is represented by quaternion $q_k=[r_w, r_x, r_y, r_z]$, and a scaling factor, represented by $s_k \in \mathbb{R}^{3}$. 
Given a 3D point $x \in \mathbb{R}^{3\times1}$, the 3D Gaussian can be formulated as: $G(x) = o \cdot e^{ -\frac{1}{2} (x-\mu)^{\top} \Sigma^{-1} (x-\mu) }$
. During the rendering process, these 3D Gaussians are splatted onto the 2D image plane. This involves a viewing transform matrix $W$ and the Jacobian matrix $J$ of the affine approximation of the projective transformation, yielding a 2D covariance matrix $\Sigma_{\text{2D}}$ through: $\Sigma_{\text{2D}} = JW\Sigma W^TJ^T$, and the 2D center position $\mu_{\text{2D}}=JW\mu$. 
The final color $C(p)$ for a pixel $p$ is computed by $\alpha$-blending the projected Gaussians, sorted by depth $C(p) = \sum_{k=1}^N SH(sh_k,v_k) \alpha_k \prod_{j=1}^{k-1} (1-\alpha_j)$, where $SH$ is the spherical harmonic function, $v_k$ is the view direction, and $\alpha_k$ represent the density computed from the $k$-th 3D Gaussian.
\paragraph{FLAG-4D}The main objective of this work is to construct a 4D reconstruction model, $F_{\omega}(\cdot)$;
\begin{multline}
    F_{\omega}(G_{0}^{i}(x, r, s, c,\sigma), t) \rightarrow  \\
    G_{t}^{i}(x+\delta x, r+\delta r, s+\delta s, c+\delta c , \sigma+\delta\sigma),
    \label{fig:vanila_dd}
\end{multline}
that takes an initial set of $n$ Gaussians, $G_{0}^{i}(x, r, s, c,\sigma)|_{i=0}^{n}$, where $x$ defines the center, $r$ is the rotation, $s$ is the scaling, $c$ is the color, and the $\sigma$ defines the opacity of a single Gaussian, and produces deformed Gaussians, $G_{t}(x+\delta x, r+\delta r, s+\delta s, c+\delta c,\sigma+\delta\sigma)$, that models the dynamics of the initial set of Gaussians $G_{0}$, by learning a deformation field by tuning the parameter, $\omega$, of the model $F_{\omega}(\cdot)$.
We initialize the set of 3D Gaussians $G_{0}$ from the sparse point cloud produced by Structure from Motion (SfM) \cite{kerbl20233d}. 
The SfM point cloud is derived from a sequence of video frames, $\{v_0, v_1, \cdots, v_T\}$ as the time varies $t = 0, \cdots, T$, captured using a monocular camera \cite{Wang_2021_CVPR}. 

\begin{figure*}
    \centering
    \includegraphics[width=\textwidth]{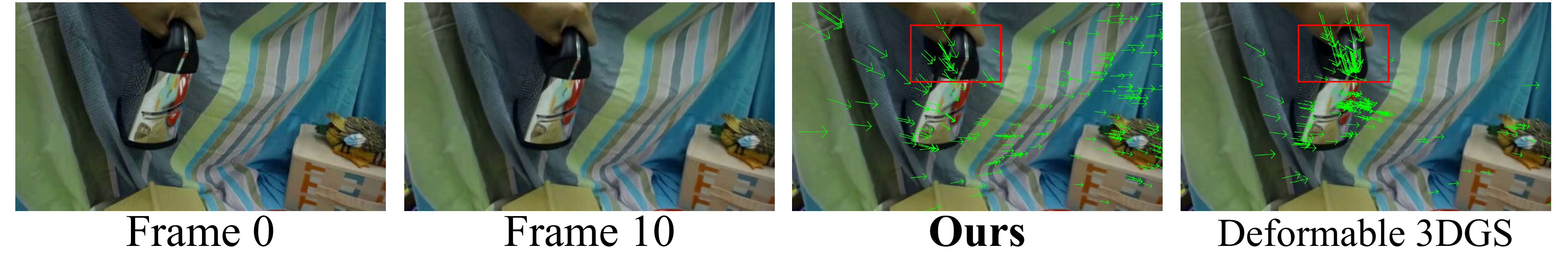}
    \caption{\textbf{Comparison of Predicted Gaussian Deformation ($t \rightarrow t+10$) for the "Bell" Scene.} (a) Frame 0 (GT). (b) Frame 10 (GT). (c) \model\ (Ours) accumulated flow from $t=0$ to $t=10$, overlaid on Frame 0. (d) Deformable 3D Gaussians \cite{yang2024deformable} accumulated flow. \model\ produces a more accurate and coherent deformation field. The highlighted region (red box) demonstrates our method's superior preservation of the bell's local rigidity during its motion towards the state in Frame 10.}\label{fig:deformation}
\end{figure*}

While complementary methodologies such as \cite{das2024neural, katsumata2023efficient, lu20243d, wu20244d, li2024spacetime} have demonstrated promising results in 4D reconstruction, a major research gap remains: these approaches largely fail to capture local consistency while preserving global dynamics. 
The \model\ employs a novel dual-deformation framework enabled with both local and global deformation to model dynamic scenes through evolving 3D Gaussian primitives. 
To this end, we adapt the standard formulation in Eqn. \ref{fig:vanila_dd} to incorporate our dual-network design:
\begin{multline}
    F_{\omega}(\ \text{IDN}(\cdot), \text{GMN}(\cdot)\ ) \rightarrow  \\
    G_{t}^{i}(x+\delta x, r+\delta r, s+\delta s,c+\delta c, \sigma+\delta\sigma)
    \label{fig:our_bd}
\end{multline}
This dual-network design is a deliberate response to the fundamental trade-off where a single network struggles to model fine-grained local dynamics and preserve global temporal coherence simultaneously. Our framework resolves this tension with Instantaneous Deformation Network, $\text{IDN}(\cdot)$, which captures high-frequency local details that monolithic models often blur, while the Global Motion Network $\text{GMN}(\cdot)$, guided by optical flow, maintains the long-range dependencies. A mutual learning strategy is used to harmonize these two network predictions. The necessity of this decomposition is quantitatively confirmed by ablation studies (Tab. \ref{tab:ablation}) and qualitatively demonstrated by the detail preservation and coherent motion in our visual results (Figs. \ref{fig:detailed-comparison},\ref{fig:deformation},\ref{fig:nerfds-quality}).

\subsection{Instantaneous Deformation for Local Deformation}
\label{secsec:idn}
Given the time embedding, $\text{Time}(t)$, and the initial set of Gaussians $G_{0}$, the IDN provides the fine-grained local deformation, see Fig. \ref{fig:bfd} bottom left, i.e.:
\begin{multline}
    \text{IDN}(\ G_{0}^{i}(x, r, s, c,\sigma), \text{Time}(t)\ ) \rightarrow \\
    G_{t, \ local}^{i}(x+\delta x, r+\delta r, s+\delta s, c+\delta c, \sigma+\delta\sigma), 
    \label{fig:idn}
\end{multline}
where $i=0,\cdots,n$.
The $\text{Time}(t)$ represents the time-driven positional encoding of the time stamp $t$ and the future time window $\{t+1, \cdots, t+W-1\}$ of window size $W$. 
The IDN takes initial Gaussians $G_{0}$ and $\text{Time}(t)$ as inputs and produces offset for initial Gaussian parameters. 

Enabling the IDN with \textit{only} time embedding, $\text{Time}(t)$ and $G_{0}$, is a necessary but not sufficient condition. 
We observed that the local deformation across time for minute objects, small colors, and discontinuous geometry requires a more informed notion of time and consistency. To this end, IDN is tasked with a two-step process.  
First, its core deformation network, $\text{IDN}_\text{core}$ is repeatedly queried for the window, \(W\), to get a more future-tailored timesteps relative to the current time \(t\) (e.g., for timesteps \(\{t+1, t+2,..., t+(W-1)\}\)). 
This yields a sequence of provisional future deformation vectors, \(\mathcal{H}_{\text{future}}^{\text{IDN}} = [\delta{\theta}_{t}^{\text{IDN}}, \dots, \delta{\theta}_{t+(W-1)}^{\text{IDN}}]\). This sequence is then processed by a Gated Recurrent Unit (GRU) \cite{dey2017gate} to distill the temporal dynamics of the hypothesized local trajectory into a single, compact representation:
\begin{equation}
    \delta{\theta}_\text{rep} = \text{GRU}(\mathcal{H}_{\text{future}}^{\text{IDN}})
    \label{eq:gru_rep}
\end{equation}
In this way, \(\delta{\theta}_\text{rep}\) encapsulates a more learned summary of the IDN's hypothesized local motion trajectory over a short future horizon. 
For clarity, we refer the querying process of the $\text{IDN}_\text{core}$ and GRU as a function of the IDN module. The GRU's role in IDN is to distill the pattern of anticipated local dynamics from each forecast and provide GMN a predictive context. Crucially, the input sequence of the GRU is generated on-the-fly by the feed-forward IDN for each frame independently. Therefore, the use of GRU does not create inter-frame dependencies and supports fast, random time access for rendering. 
\subsection{Global Motion Network for Global Deformation}
\label{secsec:gmn}
While the \model\ is refined for local motions, it is still a question of how we could leverage the global motion to model the global deformation. 
Prior works \cite{kerbl20233d, kwak2025modecgsglobaltolocalmotiondecomposition} have shown results by learning the global motion directly from SfM or by learning anchor points. 
In contrast, our GMN is designed as a multi-stage pipeline. First, a Temporal Fusion Encoder is used to process and fuse optical flow embeddings. These global motion features are then aligned with IDN's hypothesized local deformation using a Contextual Deformation Alignment. Finally, a terminal Deformation Refinement Network synthesizes these aligned features to predict the final deformation for the next timesteps.  
\noindent \textbf{Temporal Fusion Encoder.} 
We note that global motion (and hence, global deformation) can be achieved directly by leveraging optical flow. 
To explicitly capture inter-frame motion for more accurate Gaussian
deformations, the \model\ incorporates a \emph{Temporal Fusion Encoder}
that processes motion embeddings extracted from a pretrained
optical-flow network. This feature extraction is a pre-computation step for the entire video sequence, and the resulting embeddings are cached. Thus, it does not impact the final interactive rendering speed, which only relies on loading these pre-computed features.  Instead of the raw, and often noisy, 2-D
optical-flow field, we use high-dimensional intermediate embeddings
\(flow_{t-1\to t},\,flow_{t\to t+1}\!\in\!\mathbb{R}^{B\times H\times
  W\times128}\) obtained from the backbone.  
Two optical-flow embeddings—(1) past-to-current
(\(flow_{t-1\to t}\)) and (2) current-to-future
(\(flow_{t\to t+1}\))—are concatenated along the feature dimension to
form a composite tensor
\(flow_{emb}=\operatorname{Concat}[flow_{t-1\to t},\,flow_{t\to
    t+1}]\).  
The Temporal Fusion Encoder \(Fusion_{enc}\) then produces
\(
  M^{t+1}_{\mathrm{final}}
  = Fusion_{enc}(flow_{emb},\,t{+}1),
\)
which summarizes both the motion that led to the current state and the
anticipated motion of the next frame, and thus serves as the primary
image-based scene-dynamics representation.

\noindent\textbf{Contextual Deformation Alignment.}\;
Once the global motion is captured over time using the temporal fusion encoder, the remaining task is to align the global motion and the local motion encoding. 
We design a network that aligns local deformation with
global motion cues through multi-head cross-attention (MHA).  
The main purpose of the MHA is to synchronize the local deformation with the global motions, yielding temporally consistent deformation:
\begin{equation}
  M^{t+1}_{\mathrm{scene},i}
    = \mathrm{softmax}
      \!\Bigl(
        \tfrac{G_{\mathrm{unified}}M^{t+1}_{\mathrm{final}\!}{}^{\!\top}}{\sqrt{d}}
      \Bigr)
      M^{t+1}_{\mathrm{final}},
\end{equation}
where
\(G_{\mathrm{unified}}=\delta\theta_{\mathrm{rep}} +
  G^{i}\)  
is the Gaussian state across the window \([t+1,\dots,t{+}W{-}1]\).
\(G_{\mathrm{unified}}\) probes how the global motion should act, given the
Gaussian’s anticipated local evolution.  
The fused embedding \(M^{t+1}_{\mathrm{final}}\) is further enriched with sinusoidal
positional encodings to preserve spatio-temporal order and serves as
the \emph{key} and \emph{value} in MHA, while
\(G_{\mathrm{unified}}\) acts as the \emph{query}, essentially asks: ``For this primitive at this location with the anticipated local motion, what is the relevant global motion information?'' 
The result,
\(M^{t+1}_{\mathrm{scene},i}\), is an optical-flow feature that is
selectively filtered and weighted according to the IDN-predicted
future trajectory of Gaussian \(i\), giving a context-aware, dynamic
motion descriptor.

\paragraph{Deformation Refinement Network (DRN)}At \(t{+}1\), DRN forms the final predictive stage of FLAG-4D.  Acting
as an \emph{integrator}, it combines the context-aligned motion
representation \(M_{\mathrm{scene}}\) with the noise-augmented
temporal embedding \(S(t{+}1)\) to generate the refined deformation:
\begin{equation}
  \delta{\theta}_{t+1}^{\text{GMN}}
    = \operatorname{DRN}\!\bigl(\operatorname{Concat}(S(t{+}1),G_{t},\,M_{\mathrm{scene}}^{t+1}\bigr)),
\end{equation}
where \(G_{t}=G_{0}+\delta\theta_t\) and $\delta\theta_t$ is the direct forecast of $\text{IDN}_\text{core}$ at time $t$. This design ensures that the GMN's prediction is conditioned on an explicit geometric prior for the target state (IDN's direct forecast), while this prior is itself interpreted and modulated by context from both the IDN's hypothesized future trajectory and observed optical flow. 
\subsection{Mutual Learning of GMN and IDN}
\label{sec:mlearning}
We create a feedback mechanism to bring the IDN and GMN to learn from each other. This mechanism acts as a regularization in both networks, ensuring a more stable deformation prediction. The feedback mechanism is based on Mutual Learning \cite{zhang2018deepmutuallearning}, which encourages both networks to share representations and learn complementary features that are difficult to capture independently by a single network. The mutual learning objective \(\mathcal{L}_{mutual}\) is formulated to encourage agreement between their respective predictions, \(\delta\boldsymbol{\theta}_{t,i}^{\text{GMN}}\) and \(\delta\boldsymbol{\theta}_{t,i}^{\text{IDN}}\).
\begin{equation}
    \begin{split}
        \mathcal{L}_{\text{mutual}} = \frac{1}{N_{\text{vis}}} \sum_{i \in G_{vis}} \Biggl( & \left\| \text{sg}(\delta\boldsymbol{\theta}_{t,i}^{\text{IDN}}) - \delta\boldsymbol{\theta}_{t,i}^{\text{GMN}} \right\|_2^2 \\
        & + \left\| \delta\boldsymbol{\theta}_{t,i}^{\text{IDN}} - \text{sg}(\delta\boldsymbol{\theta}_{t,i}^{\text{GMN}}) \right\|_2^2 \Biggr)
    \end{split}
\end{equation} 
where \(sg(.)\) represents the stop-gradient operator. The first term treats the IDN's prediction as a pseudo-target to supervise the GMN, encouraging GMN to learn how to produce deformations that are globally consistent while still aligning with fine-grained local details. The second term treats the GMN's prediction as a pseudo-target for the IDN, allowing the IDN to learn from GMN's broader contextual understanding, which is extracted from optical flow and its look-ahead capabilities. This bidirectional distillation harmonizes their distinct inductive biases, encouraging the predictions to become consistent and complementary. 
\subsection{Optimization}
\paragraph{Smoother Opacity Regularization}
\label{secsecsec:opacity}

Managing the lifecycle of the Gaussian primitive is crucial for both quality and efficiency. Inspired by the opacity resetting strategy \cite{rota2024revising}, we design a smoother opacity regularization to promote sparsification by gradually reducing opacity. This strategy offers more stability than conventional periodic hard opacity resets, which can disrupt training and densification. The current opacity logit for a primitive $k$, $\alpha_k$, is converted to its corresponding opacity value $o_k = \sigma(\alpha_k)$, where $\sigma(\cdot)$ is the sigmoid function. We decay this opacity value $o_k$ by a small factor $\delta_o$ and clamp it within a defined range $[o_{\text{min}}, o_{\text{max}}]$ to obtain the new opacity value $o'_k$:
\begin{equation}
o'_k = \text{clamp}\left( o_k - \delta_o, o_{\text{min}}, o_{\text{max}} \right)
\label{eq:opacity_update_value}
\end{equation}
where \(\delta_o = 0.001\), and the clamping range \([o_{\text{min}}, o_{\text{max}}]\) is set to \([0.01, 1.0]\). This update occurs after each densification step. Finally, to facilitate optimization, this new opacity value \(o'_k\) is converted back to logit space, \(\alpha'_k = \text{logit}(o'_k)\), where \(\text{logit}(\cdot)\) is the inverse sigmoid function, and \(\alpha'_k\) is stored as the updated opacity logit for the primitive.

\paragraph{Depth Enhanced Loss}
\label{secsec:depth}
We obtain the normalized depth map $D_{gt}$, which serves as guidance to enhance the depth estimation. We render the depth map $D$ during the rasterization process and normalize the depth value to $[0,1]$, making the supervision robust to scale and shift differences. Averaged over all pixels $\mathcal{P}$ in the view, the loss is: 
\begin{equation}
\mathcal{L}_{\text{depth}} = \frac{1}{|\mathcal{P}|} \sum_{p \in \mathcal{P}} \left| \frac{D(p) - \min(D)}{\max(D) - \min(D)} - D_{\text{gt}}(p) \right|
\end{equation}
This encourages the rendered geometry to align with the structure provided by the external depth prior. 
\paragraph{Local Rigidity Loss}
\label{secsec:arap}
We use the local rigidity loss \cite{huang2024sc,stearns2024dynamic,yu2024cogs}, i.e. $\mathcal{L}^{\text{rigid}}$, to encourage the network to learn physically plausible motion and ensure locality in dynamic scenes. 
\begin{equation}
    \mathcal{L}^{\text{rigid}}_{i, j} = w_{i, j}||(\mu_{j, t-1} - \mu_{i, t-1}) - {R}_{i, t-1}{R}^{-1}_{i, t}(\mu_{j, t} - \mu_{i, t})||_{2},
\end{equation}
\begin{equation}
    \mathcal{L}^{\text{rigid}} = \frac{1}{k|G|}\sum_{i \in G} \sum_{j \in \text{knn}_{i;k}}\mathcal{L}^{\text{rigid}}_{i, j}.
\end{equation}
In this formulation, $\mu_{i, t}$ denotes the position of Gaussian $i$ at time $t$, with the same notation used for the time step $t-1$.
For each Gaussian $i$, neighboring Gaussian $j$ are encouraged to move in alignment with the rigid-body transform of the coordinate system of $i$ between timesteps. Here, $w_{i, j}$ represents a weighting factor, and $\text{knn}_{i;k}$ denotes the $k$-nearest neighbors of Gaussian $i$ within the Gaussians set $G$.
\begin{figure*}[t!]
    \centering
    \footnotesize{
        \setlength{\tabcolsep}{1pt} 
        \begin{tabular}{p{8.2pt}ccccc}
            & GT & Ours & SC-GS & D-MiSo & 4DGS   \\
        \raisebox{18pt}{\rotatebox[origin=c]{90}{Press}}&
         \includegraphics[width=0.18\textwidth]{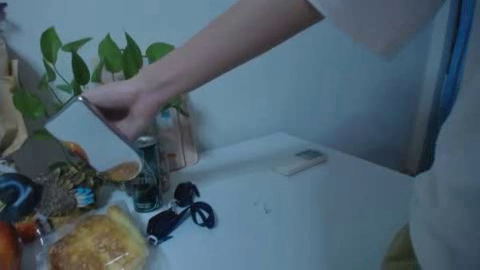} &
        \includegraphics[width=0.18\textwidth]{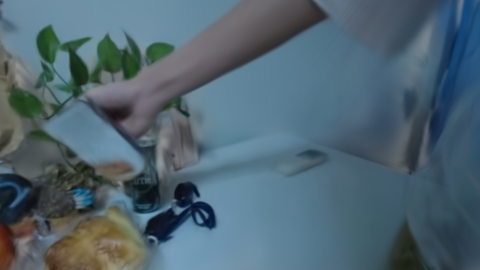} &
        \includegraphics[width=0.18\textwidth]{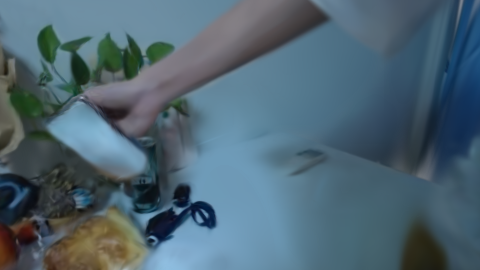} &
        \includegraphics[width=0.18\textwidth]{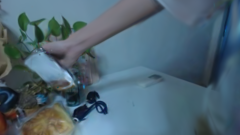} &
        \includegraphics[width=0.18\textwidth]{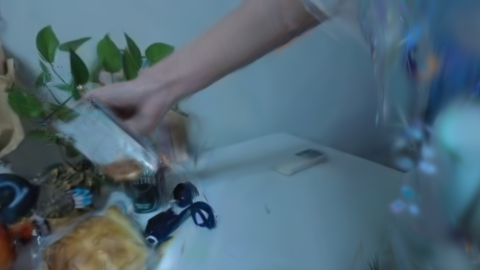}         
         \\

        \raisebox{20pt}{\rotatebox[origin=c]{90}{As}}&
         \includegraphics[width=0.18\textwidth]{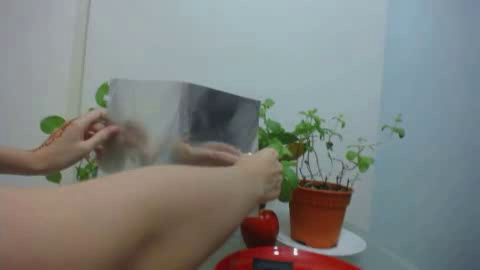} &
        \includegraphics[width=0.18\textwidth]{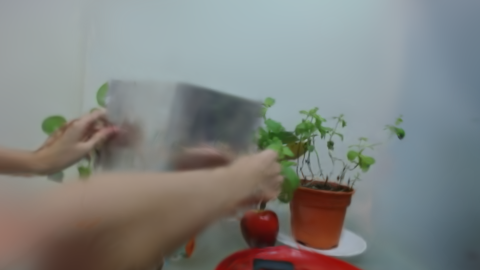} & 
        \includegraphics[width=0.18\textwidth]{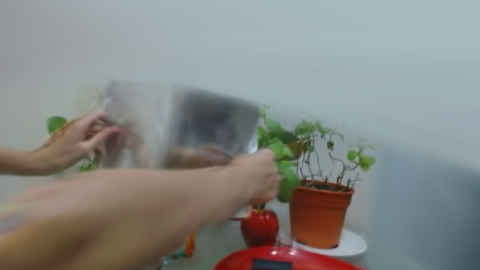} &
        \includegraphics[width=0.18\textwidth]{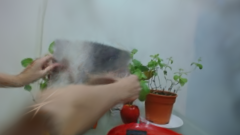} &
        \includegraphics[width=0.18\textwidth]{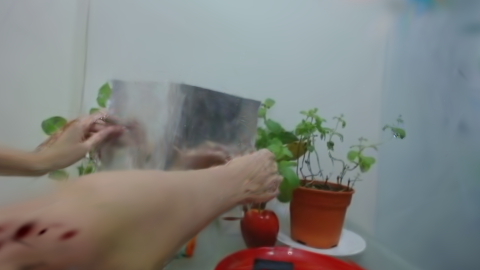} 
        \\

        \raisebox{20pt}{\rotatebox[origin=c]{90}{Sieve}}&
         \includegraphics[width=0.18\textwidth]{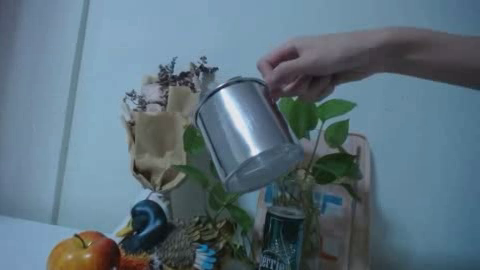} &
        \includegraphics[width=0.18\textwidth]{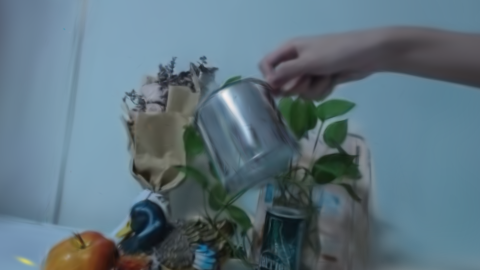} & 
        \includegraphics[width=0.18\textwidth]{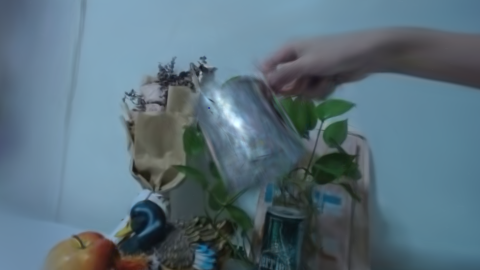} &
        \includegraphics[width=0.18\textwidth]{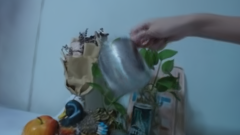} &
        \includegraphics[width=0.18\textwidth]{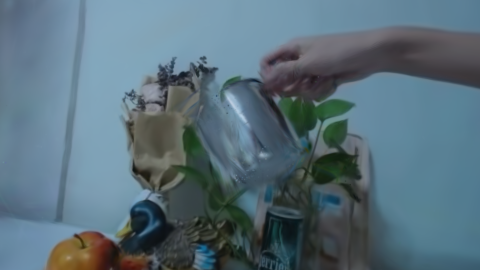} 
        \\
        \raisebox{20pt}{\rotatebox[origin=c]{90}{Bell}}&
        \includegraphics[width=0.18\textwidth]{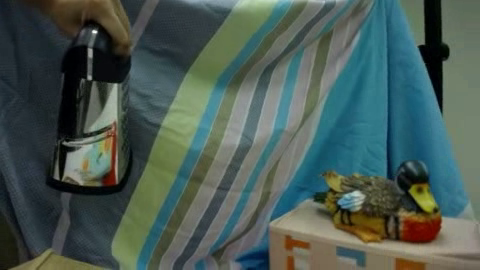} &
        \includegraphics[width=0.18\textwidth]{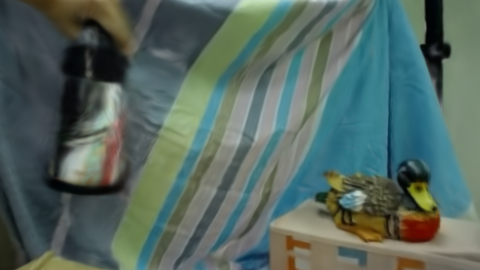} &
        \includegraphics[width=0.18\textwidth]{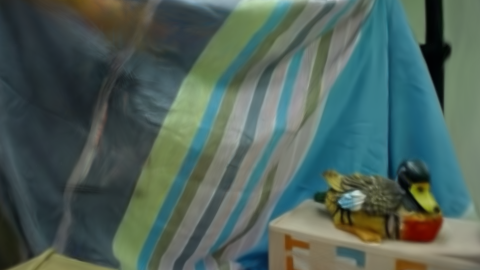} &
        \includegraphics[width=0.18\textwidth]{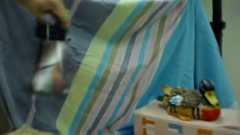} &
        \includegraphics[width=0.18\textwidth]{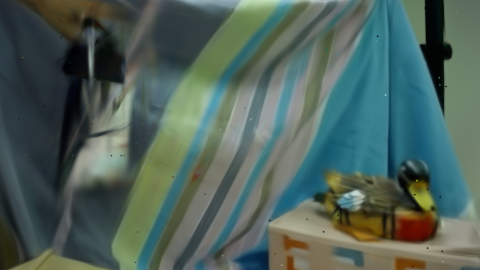} 
        \\
        \raisebox{20pt}{\rotatebox[origin=c]{90}{Cup}}&
        \includegraphics[width=0.18\textwidth]{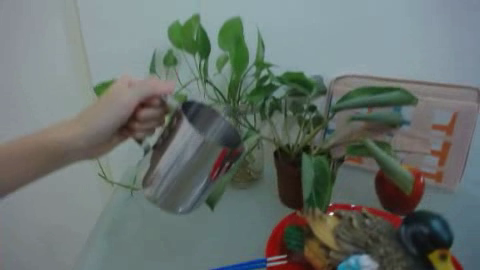} &
        \includegraphics[width=0.18\textwidth]{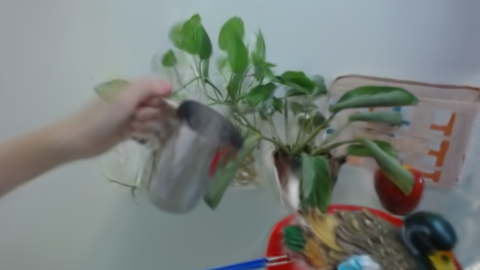} &
        \includegraphics[width=0.18\textwidth]{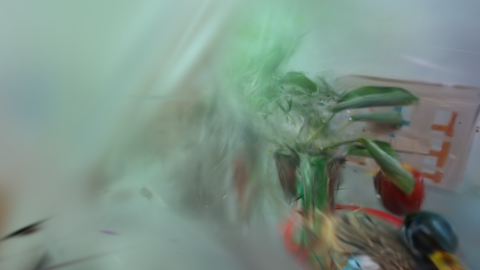} &
        \includegraphics[width=0.18\textwidth]{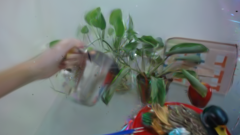} &
        \includegraphics[width=0.18\textwidth]{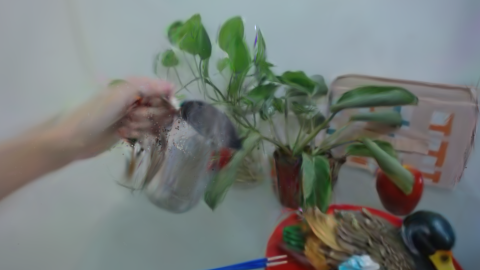} 
         \\
        \end{tabular}
    }
    \caption{\textbf{Qualitative comparisons between baseline methods and our approach on the NeRF-DS real-world dataset.} Results show that our method delivers superior rendering quality in the case of complex scene dynamics. Our method is capable of capturing finer details, preserving complex structure, and handling the dynamic scene elements more effectively than baseline methods such as SC-GS \cite{huang2024sc}, D-MiSo \cite{waczynska2024d}, and 4DGS \cite{wu20244d}. 
}\label{fig:nerfds-quality}
\end{figure*}

\begin{table}[t!]
\centering
\small
\setlength{\tabcolsep}{1.5pt} 
\begin{tabular}{l ccc}
\toprule
Method & PSNR$\uparrow$ & SSIM$\uparrow$ & LPIPS$\downarrow$ \\
\midrule
3D-GS \cite{kerbl20233d} & 20.29 & 0.782 & 0.292 \\
TiNeuVox \cite{fang2022fast} & 21.61 & 0.823 & 0.277 \\
HyperNeRF \cite{park2021hypernerf} & 23.45 & 0.849 & 0.199 \\
NeRF-DS \cite{yan2023nerf} & 23.60 & 0.849 & 0.182 \\
Deformable 3DGS \cite{yang2024deformable} & 23.76 & 0.848& 0.180 \\
SC-GS \cite{huang2024sc} & 22.25 & 0.824 & 0.203 \\
D-MiSo \cite{waczynska2024d} & 23.90 & 0.851 & \textbf{0.151} \\
4DGS \cite{wu20244d} & 22.54 & 0.837 & 0.212 \\
MotionGS \cite{zhu2024motiongs} & 23.71 & 0.831 & 0.240 \\
\midrule
Ours & \textbf{24.23}  & \textbf{0.852} & 0.198 \\
\bottomrule
\end{tabular}
\caption{Quantitative comparison on NeRF-DS \cite{yan2023nerf} dataset. Mean performance across all scenes.}
\label{tab: nerfds-comparison}
\end{table}

\begin{table}[t!]
\centering
\small
\setlength{\tabcolsep}{4pt} 
\begin{tabular}{l cc}  
\toprule  
Method & PSNR$\uparrow$ & SSIM$\uparrow$ \\
\midrule
SC-GS~\cite{huang2024sc} & 20.92 & 0.63 \\
D-MiSo~\cite{waczynska2024d} & \textbf{22.47} & 0.61 \\  
Deformable 3DGS~\cite{yang2024deformable} & 22.06 & 0.58 \\
MotionGS~\cite{zhu2024motiongs} & 20.96 & 0.49 \\
\midrule
Ours & 22.33 & \textbf{0.78} \\
\bottomrule  
\end{tabular}  
\caption{Quantitative comparison on HyperNeRF's \cite{park2021hypernerf} vrig dataset. Mean performance across all scenes.} 
\label{tab: hypernerf_comparison} 
\end{table}
\paragraph{Total Loss}
The total loss can be summarized as: 
\begin{equation}
    \mathcal{L} = \mathcal{L}_\text{mutual}+\mathcal{L}_\text{depth}+\mathcal{L}_\text{rigid}+\mathcal{L}_\text{render} 
\end{equation}
representing the mutual learning loss, depth-enhanced loss, local rigidity loss, and rendering loss. The rendering loss is a combination of $L_1$ loss and D-SSIM loss, which balances sharpness and structural coherence. 
\section{Experiments}
\label{sec:experiments}
\begin{figure}
    \centering
    \includegraphics[width=\columnwidth]{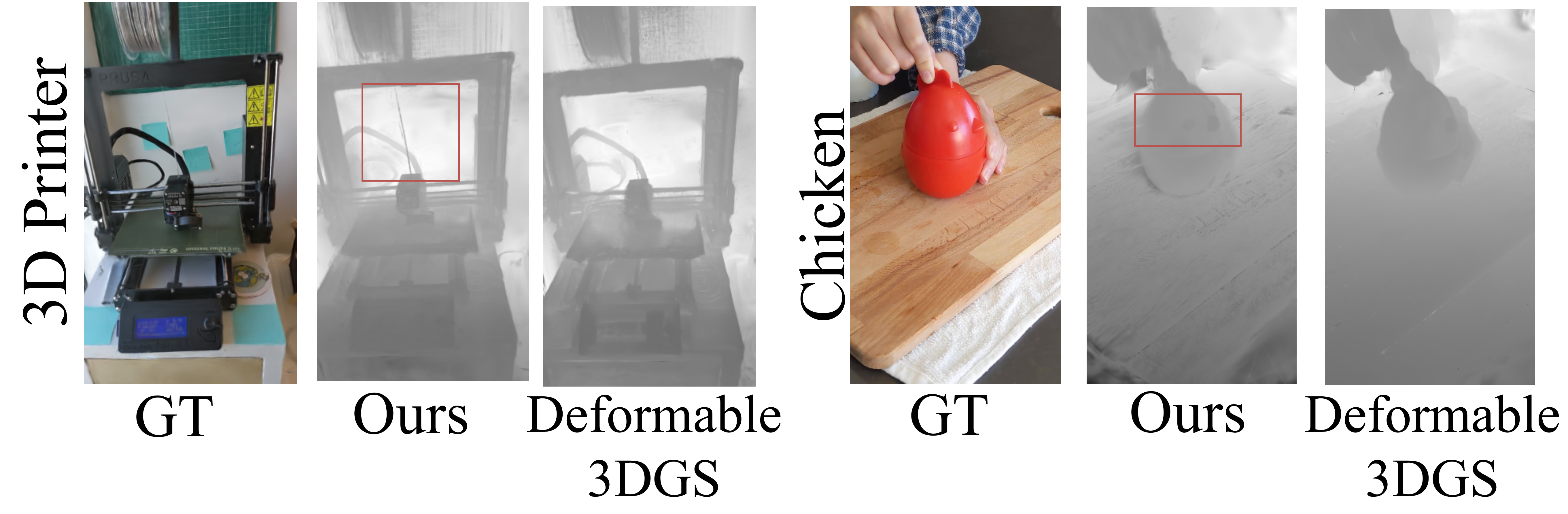}
    \caption{\textbf{Rendered Depth Map Comparison} \model\ produces depth maps with visibly greater detail and geometric accuracy (e.g., printer's string and surface texture). 
    }\label{fig:hypernerf-depth}
\end{figure}
\begin{figure}[ht]
\centering
    \includegraphics[width=\columnwidth]{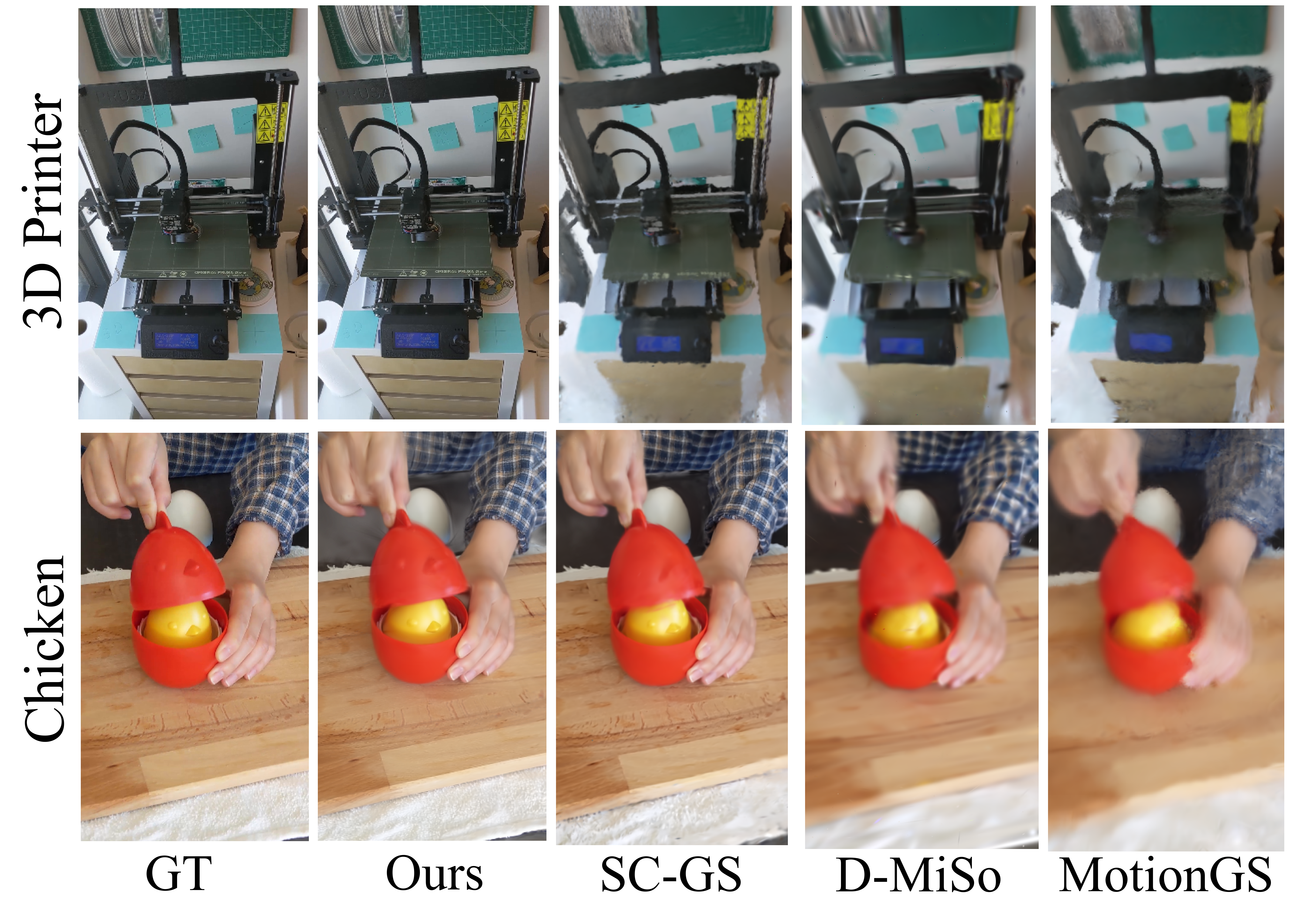}
    \caption{\textbf{(Best viewed while zoomed-in.)} Qualitative comparisons between baselines and our approach on the HyperNeRF dataset. The comparison illustrates our model achieves higher rendering quality consistently in challenging real-world scenes where camera pose estimation is less accurate.}\label{fig:hypernerf-comparison}
\end{figure}
In the experiment setup, we use GeoWizard \cite{fu2024geowizard} as our depth estimation model to guide the depth-enhanced loss, and MemFlow \cite{dong2024memflow} is used as the pre-trained optical flow network to provide guidance. 
The experiment is conducted on a single A100 GPU. For all datasets, we use Adam optimizer \cite{kingma2014adam} to optimize both the IDN and GMN, along with the 3D Gaussian parameters end-to-end. The learning rate for IDN and GMN parameters is set to $1e-4$. A time window of 4 is used for IDN's future hypothesis sequence.  
To evaluate the performance of our method, we perform experiments on a monocular real-world dataset from NeRF-DS \cite{yan2023nerf} and HyperNeRF \cite{park2021hypernerf}. We utilize standard image quality metrics, PSNR, SSIM, and LPIPS, to assess the effectiveness of our method. 
\subsection{Results}
\paragraph{Comparisons on NeRF-DS dataset.} 
NeRF-DS dataset is a challenging benchmark that contains complex scene structures and motion. Tab. \ref{tab: nerfds-comparison} shows that our approach achieves state-of-the-art (SOTA) performance across a majority of these scenes, indicating high reconstruction fidelity. We note a favorable trade-off in the LPIPS perceptual metric, where our framework's deliberate focus on preserving sharp details and eliminating motion blur results in a competitive score while excelling in PSNR and SSIM \cite{zheng2024structure}. The qualitative results in Fig. \ref{fig:nerfds-quality} highlight that our model is more effective in capturing finer details and maintaining clarity. This enhanced coherence is not only aesthetic, but is also evident in the underlying learned motion. As Fig. \ref{fig:deformation} directly visualizes, the accumulated deformation field generated by our framework is more structured and physically plausible than the baseline method. It preserves the object's rigidity correctly over a long time horizon, while the baseline produces a less coherent flow. 
\paragraph{Comparisons on HyperNeRF dataset.}
Tab. \ref{tab: hypernerf_comparison} summarizes the quantitative performance on HyperNeRF. We obtain SOTA performance, highlighted by the strongest SSIM scores in every scene. The qualitative comparison of HyperNeRF is shown in Fig. \ref{fig:hypernerf-comparison}. Our approach produces results with the best sharpness and visual quality among the other methods. The zoom-in comparison is shown in Fig. \ref{fig:detailed-comparison}. While other methods have blurred or illegible text, our method preserves the fine textual details of the wording on the machine more accurately, with greater sharpness and clarity. 
\subsection{Ablation Studies}
\paragraph{Removing IDN} 
In this ablation, we evaluate a variant of our model where the IDN is removed ("FLAG-4D w/o IDN"). This setup relies solely on the GMN for the deformation prediction. GMN still receives temporal signals and fused optical flow embeddings. However, the attention mechanism and final prediction lack the local deformation hypotheses and hypothesized future local trajectory $\delta\theta_{rep}$, which are provided by IDN. From Tab. \ref{tab:ablation}, the performance drops even when optical flow is provided to GMN, and it struggles to reconstruct these high-frequency spatial and temporal details.
\paragraph{Removing GMN}
Conversely, we evaluate another variant where the GMN is disabled ("FLAG-4D w/o GMN"). In this setup, the deformations are predicted solely by the IDN, based on its temporal signal $\text{Time}(t)$ and its internal GRU processing its own future hypothesized local trajectory. This variant lacks the integration of external optical flow features and the global refinement provided by the GMN's attention mechanism. Disabling GMN also results in a degradation in performance, indicating that it struggles with long-range temporal coherence and global consistency. 
\paragraph{Disabling Mutual Learning}
IDN and GMN are trained separately by their own losses. Disabling Mutual Learning results in degradation in performance; this is attributed to the fact that the IDN and GMN can diverge in their learned representation. Mutual Learning acts as an essential regularizer, encouraging GMN to respect local details while forcing the IDN to adhere to global motion. The degradation further proves that mutual learning is essential for integrating the strengths of local and global networks into a unified, high-fidelity representation.

\begin{table}[t]
  \centering
  \setlength\tabcolsep{0pt}
    \begin{tabular*}{\linewidth}{@{\extracolsep{\fill}} l c c c}
        \toprule[1pt]{\footnotesize{Methods}} & {\footnotesize{PSNR($\uparrow$)}} & {\footnotesize{M-SSIM($\uparrow$)}} & {\footnotesize{LPIPS($\downarrow$)}} \\
          \midrule
        \footnotesize{\model\ w/o IDN}
        & \footnotesize{24.01} & \footnotesize{0.868} & \footnotesize{0.226}      \\
        \footnotesize{\model\ w/o GMN}
        & \footnotesize{23.39} & \footnotesize{0.853} & \footnotesize{0.223}      \\
        \footnotesize{\model\ w/o Mutual Learning}
        & \footnotesize{23.70} & \footnotesize{0.878} & \footnotesize{\textbf{0.157}}        \\
        \footnotesize{\model\ }
        & \footnotesize{\textbf{24.23}} & \footnotesize{\textbf{0.884}} & \footnotesize{\textbf{0.157}}     \\

        \bottomrule[1pt]
    
      \end{tabular*}
      \caption{Ablation studies. We evaluate the effect of the proposed IDN, GMN, and the Mutual Learning strategy on the NeRF-DS dataset \cite{yan2023nerf}.}
  \label{tab:ablation}
\end{table}
\section{Conclusion}
\label{sec:conclusion}
In this work, we introduce FLAG-4D, a dual-network framework that harmonizes the core tension between local detail and global coherence in 4D reconstruction. Our framework delegates these tasks to two synergistic specialists, IDN for local detail and GMN for global coherence. These specialists are tightly coupled by our CDA mechanism. By directly leveraging the optical flow embeddings as input to guide the deformation, our method achieves state-of-the-art performance, delivering high-fidelity results with strong temporal consistency. Future work can focus on reducing the dependency on pre-trained flow networks to improve robustness in scenes with significant motion blur, where flow estimation can easily fail. 
\section{Acknowledgments}
\label{sec:acknowledgement}
We acknowledge the GRS support from the School of Information Technology, as well as the support and provision of GPU resources from the HPC/APC team and its manager, Dr. Marcus. Dr. Srinivas and Dr. Arghya acknowledge the support of the Anusandhan National Research Foundation (ANRF), Government of India (CRD/2024/000973).
\bibliography{aaai2026}

\end{document}